\documentclass[runningheads]{llncs}
\usepackage{graphicx}
\usepackage{amsmath,amssymb,amsfonts}
\usepackage{xcolor}
\usepackage[noadjust]{cite}

\begin{document}
\title{On Effects of Compression with Hyperdimensional Computing in Distributed Randomized Neural Networks\thanks{
The work of DK was supported by the European Union's Horizon 2020 Programme under the Marie Skłodowska-Curie Individual Fellowship Grant (839179) and in part by the DARPA's AIE (HyDDENN) program and by AFOSR FA9550-19-1-0241.
}
}
%
%
\author{Antonello Rosato\inst{1}\and
Massimo Panella\inst{1}\and
Evgeny Osipov\inst{2}\and
Denis Kleyko\inst{3,4}}
\authorrunning{A. Rosato et al.}
%
\institute{University of Rome ``La Sapienza'', Rome, Italy 
\email{\{antonello.rosato,massimo.panella\}@uniroma1.it}\\ 
\and Lule\aa{} University of Technology, Lule\aa{}, Sweden; \email{evgeny.osipov@ltu.se}
\and Research Institutes of Sweden, Kista, Sweden; 
\email{denis.kleyko@ri.se}
\and University of California, Berkeley, USA
}
\maketitle              
\begin{abstract}
A change of the prevalent supervised learning techniques is foreseeable in the near future: from the complex, computational expensive algorithms to more flexible and elementary training ones. 
The strong revitalization of randomized algorithms can be framed in this prospect steering.
We recently proposed a model for distributed classification based on randomized neural networks and hyperdimensional computing, which takes into account cost of information exchange between agents using compression. 
The use of compression is important as it addresses the issues related to the communication bottleneck, however, the original approach is rigid in the way the compression is used. 
Therefore, in this work, we propose a more flexible approach to compression and compare it to conventional compression algorithms, dimensionality reduction, and quantization techniques. 
\end{abstract}

%
%
%

\section{Introduction}

In this work, we are exploring the use of compression in a framework for distributed classification. 
The main motivation of this work is the optimization of the distributed machine learning algorithm when computational and communication costs are to be preserved as is the case in, e.g., resource-constrained scenarios (e.g., edge machine learning~\cite{YaziciEdge2018}, smart sensing).
Broadly, the objective for compression can be formulated as finding the best trade-off between classification performance and the communication overhead, which is crucial in the distributed scenario.  
In this scenario, and generally in  distributed solutions, the communication bottleneck is often overlooked, since the tendency is to focus on efficient algorithmic choices and implementations improving classification performance. 
We, however, aim at developing a flexible solution, which allows controlling the trade-off between classification performance and reduction of the communication overheads.

In ~\cite{RosatoHDDistributed2021} we presented a novel concept for distributed learning based on a class of randomized neural networks known as Random Vector Functional Link (RVFL) networks~\cite{RVFLorig, ELM06, Scardapane2017} using framework of hyperdimensional computing also known as vector symbolic architectures (HDC/VSA)~\cite{PlateHolographic2003, RachkovskijStructures2001,MAP,Kanerva2009, FradySDR2020}.

The main contribution of  this article is the in-depth empirical evaluation of the effect of compressing information shared between agents participating in the distributed learning. We present the trade-offs between the accuracy of the neural model and various compression approaches as means of reducing the risk  for the communication bottleneck. These results further extend the functionality of our distributed RVFL learning solution with a generalized approach for controlling the compression ratio under different communication bottleneck conditions.

The experiments reported in this paper were done with two approaches to forming network's classifier: regularized least squares (RLS) and centroids. 
Though, other alternatives are available~\cite{Yeseong2018, ShridharEnd2End2020, DiaoGLVQHD2021, HyperEmbed}.
The activations of the hidden layer of the network were formed according to a recently proposed version of RVFL network~\cite{intRVFL2020}, which simplifies the conventional architecture~\cite{RVFLorig} using some of the HDC/VSA principles~\cite{intESN2020, intSOM}.



When studying the effect of compressing the classifier, we focused on analyzing our inherently lossy compression approach in a range of  compression ratios, along with the conventional, entropic lossless compression. 
It was also important to contrast the proposed approach with some other conventional lossy approaches. 
However, we have not identified such an approach, which would compress generic numeric matrices (i.e., classifiers) without assuming a particular data modality (e.g., images) since for such data lossy compression is well performed based on perceptual (e.g., visual) models.
However, a lossy compression of the matrix containing the classifier can be done with the conventional methods for dimensionality reduction. 
In particular, we implemented a lossy approach, which relies on the eigenvectors and eigenvalues obtained from the Singular Value Decomposition (SVD) of the classifier. 
The approach uses the fact that some of the  eigenvalues with the smallest values and the corresponding eigenvectors can be ignored when reconstructing the classifier.

\section{Compression algorithms}

As already stated, the main purpose of the compression in the considered distributed scenario is in reducing the communication overhead, thus, improving the feasibility and applicability of the distributed classification. 
We are particularly interested in the lossy compression, which might results in decreased classification performance compared to the uncompressed solution, since it should allow controlling the trade-off between the compression ratio and classification performance losses. 

To better understand the experiments reported hereafter, in which we assess and analyze the performance of the proposed approach, it is important to clarify the parts of the model involved in the compression.
The network consists of connected agents (we assume full connectivity in our experiments); each agent has its own training dataset.
We form agents' datasets by equally splitting samples of the whole dataset between agents without replacement.
So each agent has the access to a subset of samples of the whole dataset with all the features. 
Each agent trains its own local model with its portion of the dataset and shares information about its own classifier (denoted as $\mathbf{W}^{\mathrm{out}}$) with all other agents (due to the full connectivity), thus, the training data are not actually shared but kept private. 
Additionally, in this work, before sharing values of its local classifier with others, each agent compresses it.
At the receiving end, each agent decompresses the data it received and incorporates them in the aggregated model to be able to create a more powerful classifier.

Below, we describe three compression approaches used herein highlighting their differences. 
Classification performance obtained with these approaches is reported and discussed in Section \ref{sec:results}.

\subsection{Compression based on conventional algorithms}
We describe hereafter the two compression approaches used to compare with the proposed approach. 
As already stated, they are general and, although, implemented specifically for the purpose of the presented distributed classification, they only make use of well-known conventional algorithms.

\subsubsection{Lossless compression}
The most common and effective way to perform compression on a matrix is using one of the many conventional lossless  compression algorithms readily available and widely assessed.
We opted for the ones in the \texttt{zlib} software library for data compression, which use the public domain ZLIB Deflator algorithm. 
The algorithm uses the Deflate file format, employing a combination of LZSS and Huffman coding (see~\cite{10.17487/RFC1950} for more details).
In this work, the lossless compression is given as a baseline of how strong is the gain of the proposed approach  in terms of information compression. 
Since it is impractical from the application point of view, given that the operations needed to perform it at the local nodes are computationally expensive with respect to the actual classification model.
Moreover, when using the lossless compression, the classification performance of the model is the same as in the uncompressed case but the compression ratio is fixed to the one obtained by the compression algorithm.

\subsubsection{Lossy compression}
\label{sec:comp:SVD}
Lossy quantization of a data matrix is well performed in the case of audio and video applications, using specific techniques.
Such techniques are not of interest herein since they are based on auditory and visual perceptual models that are not applicable in our scenario. 
Nevertheless, one way to decrease the size of the classifier is by quantizing matrix values to some fixed number of quantization levels.  
We will discuss this apporoach in Section~\ref{subsec:quant}. 


In the absence of quantization, dimensionality reduction methods can serve for the purpose of compression. 
In particular, we use the eigenvectors and eigenvalues obtained from the SVD and the compression is based on the fact that eigenvalues with smaller, negligible norm can be discarded. 

In practice, we carry out the SVD-based  compression by first transforming the classifier $\mathbf{W}^{\mathrm{out}}$ with the SVD. 
Given that the values of $\mathbf{W}^{\mathrm{out}}$ are real, this results in the well-known factorization: $\mathbf{U \Sigma V}$. 
Given the properties of the SVD, if the eigenvalues and both the left and right eigenvectors are transmitted as is, another agent is able to losslessly reconstruct the original matrix.
Of course, this is completely impractical for compression and sharing, since the dimension of the transmitted information in the form of the three matrices $\mathbf{U}$, $\mathbf{\Sigma}$ and $\mathbf{V}$ is much larger than the original classifier $\mathbf{W}^{\mathrm{out}}$.

For the compression purpose, to ensure the reduction of the size of the matrices to be transmitted, each agent selects and transmits only a portion of the SVD matrices, corresponding to a number $t$ of the largest eigenvalues, resulting in what is often called a ``truncated SVD'' ($\mathbf{U}_t$, $\mathbf{\Sigma}_t$ and $\mathbf{V}_t$), as it is used for proper dimensionality reduction~\cite{xu1998truncated}.
This way, the receiving agent is able to reconstruct a version of the original classifier: $\hat{\mathbf{W}}^{\mathrm{out}}=\mathbf{U}_t \mathbf{\Sigma}_t \mathbf{V}_t$, which is not an exact replica, since some information is lost due to the truncation.
To be able to effectively control the lossy compression ratio of the classifier, we reshape $\mathbf{W}^{\mathrm{out}}$ into a square matrix (with zero padding if necessary) and then select $t$ eigenvalues based on the desired compression ratio.


\subsection{Compression based on HDC/VSA principles}
\label{sec:comp:HD}


In our previous work~\cite{RosatoHDDistributed2021}, we proposed how to use the principles of HDC/VSA to compress the classifier $\mathbf{W}^{\mathrm{out}}$. 
Here we recapitulate it. 
The key idea is that before being shared with the other agents a locally computed version of $\mathbf{W}^{\mathrm{out}(p)}$ can be compressed into a single $D$-dimensional vector (hypervector, denoted as $\mathbf{w}$)
For the proposed approach we use Holographic Reduced Representations model~\cite{PlateHolographic2003}.

The compression procedure uses the idea of key-value pair representations in HDC/VSA\footnote{
This is possible since HDC/VSA provide primitives for representing structured data in hypervectors such as sequences~\cite{Kanerva2009, HannaganHolographic2011, Frady17, ThomasHDFoundations2020}, sets~\cite{Kanerva2009}, \cite{KleykoABF2020}, state automata~\cite{YerxaUCBHD_FSA2018}, \cite{OsipovHD_FSA2017}, hierarchies, or predicate relations~\cite{PlateHolographic2003}, \cite{RachkovskijStructures2001}. 
Please consult~\cite{KleykoComputingParadigm2021} for a general overview.
}.
In the compression context, a key hypervector is generated randomly where a value hypervector contains some of the values from $\mathbf{W}^{\mathrm{out}(p)}$. 
The hypervector of a key-value pair is formed via the binding operation. 
In the Holographic Reduced Representations model, the binding operation is realized via the circular convolution (denoted as $\circledast$), which is formulated through the outer product of the hypervectors being bound.
The value of the $j$th component of $\mathbf{z}$ is calculated as: 
\noindent
\begin{equation*}
\mathbf{z}_j= \sum_{k=0}^{D-1}  \mathbf{y}_k \mathbf{x}_{j-k \mod D}
\end{equation*}
\noindent
In~\cite{RosatoHDDistributed2021}, it was proposed to form $L$ key-value pairs where $L$ denotes the number of classes in the task. 
The disadvantage of this proposal is that the compression ratio is fixed to $L$ to so the procedure is inflexible.
Here we show that the procedure can be simply modified so that the compression ratio can take any positive integer value. 
To do so, for the chosen compression ratio (denoted as $R$), we first calculate the number of dimensions $D$ in $\mathbf{w}$ as:
$D= \lceil HL/R \rceil$,
where $H$ denotes the size of the hidden layer.
Then $\mathbf{W}^{\mathrm{out}}$ should be reshaped (denoted as $\mathbf{S}$) such that is has $D$ rows and $R$ columns. 
Note that some zero padding might be necessary when $HL$ is not a multiple of $R$.
Next, $R$ random hypervectors (denoted as $\mathbf{K}_i$) are generated.
They act as keys. 
Each column in $\mathbf{S}$ is then bound with the corresponding key:  $\mathbf{K}_i \circledast \mathbf{S}_i$.
Finally, these key-value hypervectors are used to form the compressed version of the classifier: 
\noindent
\begin{equation*}
\mathbf{w}= \sum_{i=1}^R \mathbf{K}_i \circledast \mathbf{S}_i;
\end{equation*}
\noindent
$\mathbf{w}$ can be shared with other agents in an attempt to improve the classification.

When the agent $p$ receives $\mathbf{w}$ from its neighbor $q$ it needs to decompress $\mathbf{W}^{\mathrm{out}(q)}$ from $\mathbf{w}$.
The decompression is done for each $\mathbf{S}^{(q)}_i$ using the inverses of the corresponding key hypervectors of $q$ (see~\cite{PlateHolographic2003} for the details) as:
\noindent
\begin{equation*}
\hat{\mathbf{S}}^{(q)}_i \approx \mathbf{w}  \circledast \mathbf{K}_i^{-1}.
\end{equation*}
\noindent
Finally, $\hat{\mathbf{S}}^{(q)}$ has to be reshaped to get $\hat{\mathbf{W}}^{\mathrm{out}(q)}$.
Note that  the reconstructed classifier $\hat{\mathbf{W}}^{\mathrm{out}(q)}$ will not be the exact replica of the original one. 
This is explained by the fact that the superposition operation is lossy in a sense that hypervectors of other key-value pairs add their crosstalk noise to the resultant hypervector~\cite{Frady17, KleykoPerceptron2020}. 
It was, however, shown in~\cite{RosatoHDDistributed2021} that when many reconstructed classifiers are combined, the noise will average out without affecting much the classification performance. 
At the same time, it is expected that the compression ratio $R$ will regulate how lossy is the compression and, therefore, it will affect the classification performance. 
The effect of $R$ on the classification performance is shown experimentally in Section~\ref{sec:results}.


\section{Experiments}

\subsection{Data and setup}
\label{sec:data}

The experiments reported in this paper were performed on a collection of $18$ real-world classifications datasets from the UCI Machine Learning Repository~\cite{Dua2019}. 
The datasets are a subset of a larger collection used in the seminal work~\cite{Delgado2014}.
The only addition, which was introduced to the preprocessing, was the normalization of input features to $[0, 1]$ range as this is necessary to form the thermometer codes used in the model.
In the distributed scenario, it is important that each agent gets some training samples. 
Since we limited ourselves to splitting training data between agents without the replacement, only datasets with more than $1,000$ samples in the training part were considered. 
Another requirement was the size of the hidden layer, $H$.
In order to be able to make a comparison between compressed classifiers and smaller models of the corresponding size, only the datasets where the optimal $H$ was more than $1,300$ were picked. 
Only $18$ datasets in the collection satisfied both requirements\footnote{
The names of the datasets were:
Abalone,
Cardiotocography (10 classes)
Chess (King-Rook vs. King-Pawn),
Letter,
Oocytes merluccius nucleus 4D, 
Oocytes merluccius states 2F,
Pendigits,
Plant margin,
Plant shape,
Plant texture,
Ringnorm,
Semeion,
Spambase,
Statlog landsat,
Steel plates,
Waveform,
Waveform noise,
Yeast.
}. 

The optimal hyperparameters of the models ($H$, $\lambda$, and $\kappa$) for each dataset  were obtained  with the grid search using the same steps as in~\cite{Delgado2014} using the centralized scenario with the RLS classifier.
$H$ varied in the range $[50, 1500]$ with step $50$;  $\lambda$ varied in the range $2^{[-10, 5]}$ with step $1$; and $\kappa$ varied between $\{1,3,7,15\}$.
The chosen values were used for both classifiers and in all the experiments reported below. 
To avoid the influence of a particular random initialization on the performance, all results reported below were averaged for $10$ random initializations of $\mathbf{W}^{\mathrm{in}}$, which is the input projection matrix.

\subsection{Results}
\label{sec:results}

\begin{figure}[t]
    \centering
    \includegraphics[width=1.0\columnwidth]{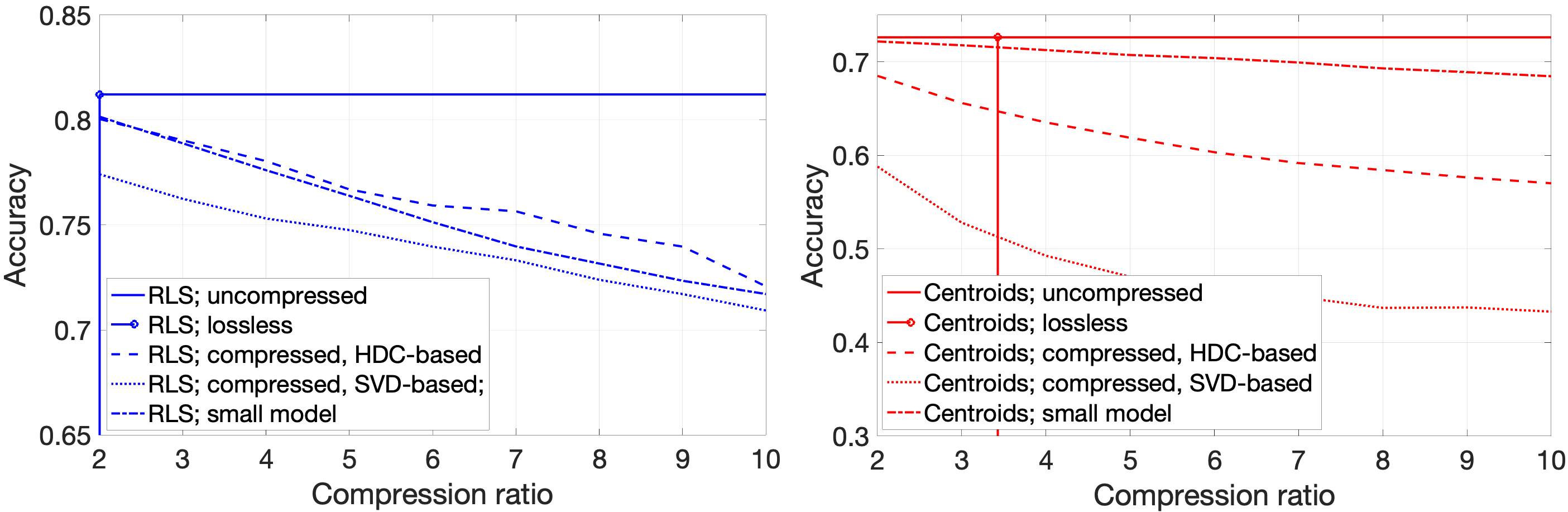}
    \caption{The average accuracy on the datasets against different compression ratios in the proposed procedure.
    The number of agents was set to $10$.
    }
    \label{fig:exp:comp:10}
\end{figure}

\begin{figure}[t]
    \centering
    \includegraphics[width=1.0\columnwidth]{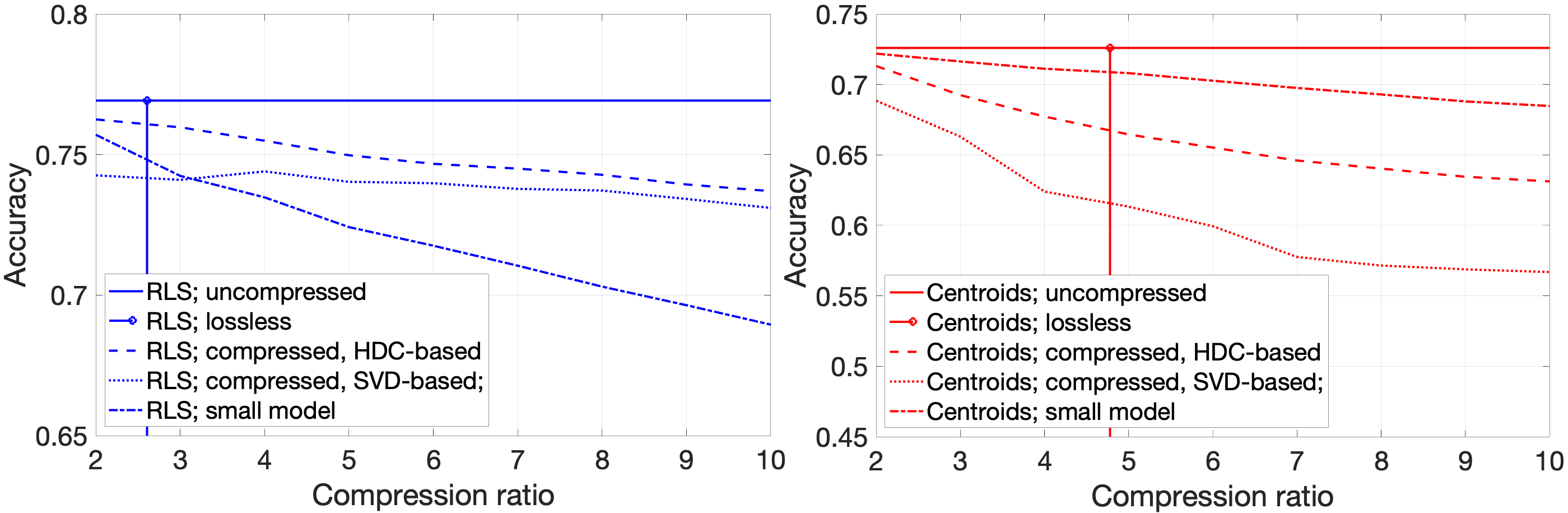}
    \caption{The average accuracy on the datasets against different compression ratios in the proposed procedure.
    The number of agents was set to $100$.
    }
    \label{fig:exp:comp:100}
\end{figure}

For the lossless compression, the results were analyzed only in terms of a single compression ratio. 
This is straightforward because, being lossless, no information is missing at the decompression receiving end of the network, and each agent can retrieve the exact copy of the classifier sent by its neighbors.
Thus, in Fig.~\ref{fig:exp:comp:10} for $N=10$ agents and Fig.~\ref{fig:exp:comp:100} for $N=100$ agents, the achieved compression ratio for the lossless algorithm is depicted by bars. 
Naturally, the accuracies were equal to the corresponding uncompressed versions of the classifiers.
We stress the fact that this case has only the effect of making the lossy results be examined in terms of compression ratio, giving us the comparison with a compression procedure optimized for much larger models and agents with superior computational power.

The advantage of the lossless compression is that it preserves all of the accuracy but there is no way to control its compression ratio.  
The proposed compression procedure (Section~\ref{sec:comp:HD}) and the SVD-based one (Section~\ref{sec:comp:SVD}) are lossy but they allow varying the compression ratio.  

In order to investigate the effect of the compression ratio on the accuracy, we performed the experiments in the distributed scenario with both procedures for both classifiers. 
Fig.~\ref{fig:exp:comp:10} presents the results of the experiments for the case when the number of agents was set $N=10$ while Fig.~\ref{fig:exp:comp:100} reports the results for $N=100$. 
As the baseline, the figures depict the accuracies of the distributed scenario when no compression was evolved (solid lines). 
For both classifiers, this baseline acted as an upper bound as losses introduced to the classifiers during the decompression incurred decreased classification performance. 
Obviously, the accuracy of both classifiers was getting worse with the increased compression ratio but given the same ratio, the RLS classifier always outperformed the centroids one.

Note that an approach alternative to the compression would be to train a smaller model and share it with other agents without involving any compression. 
Therefore, for each compression ratio we trained a model with smaller hidden layer (denoted as ``small model'' in the figure) such that the size of smaller model's classifier would equal to the compressed classifier of the full-size model. 
In the case of the centroids classifier for both $N$, small models (dash-dotted line) performed much closer to the baseline than the compressed centroids. 
The explanation is likely rooted in the fact that to have a fair comparison in the experiments we used the size of the hidden layer optimized for the RLS classifier. 
Since the centroids classifier is much simpler it is likely that it would need smaller size of the hidden layer to get most of its classification performance and so its small models were pretty close to the baseline.  
In contrast, for the RLS classifier and the proposed approach when $N=10$, for small compression ratios ($2$ or $3$) the compressed classifier and small model performed on a par. 
However, when the compression ratio was increasing, the compressed classifier was performing better than the corresponding small model. 
In the case of $N=100$, the proposed approach was noticeably better than the small models. 
That is expected because in the proposed approach noise introduced to the decompressed classifiers gets averaged out and the more classifiers get aggregated the better it is for mitigating the noise. 

With respect to the SVD-based compression, for both values of $N$, the trend of results in terms of accuracy/compression ratio trade-off was as expected.
In particular, the mean values follow the same trend as for the proposed approach, with the difference that the achieved accuracy was always lower. 
This can be likely caused by the fact that even least significant eigenvectors bear important information about the classifier so when truncated, the decompressed classifier misses important information.
Results indicated that SVD-based compression, while remaining a solid dimensionality reduction method, was inferior in terms of accuracy/compression ratio trade-off compared to the proposed approach.

\begin{figure}[t]
    \centering
    \includegraphics[width=0.88\columnwidth]{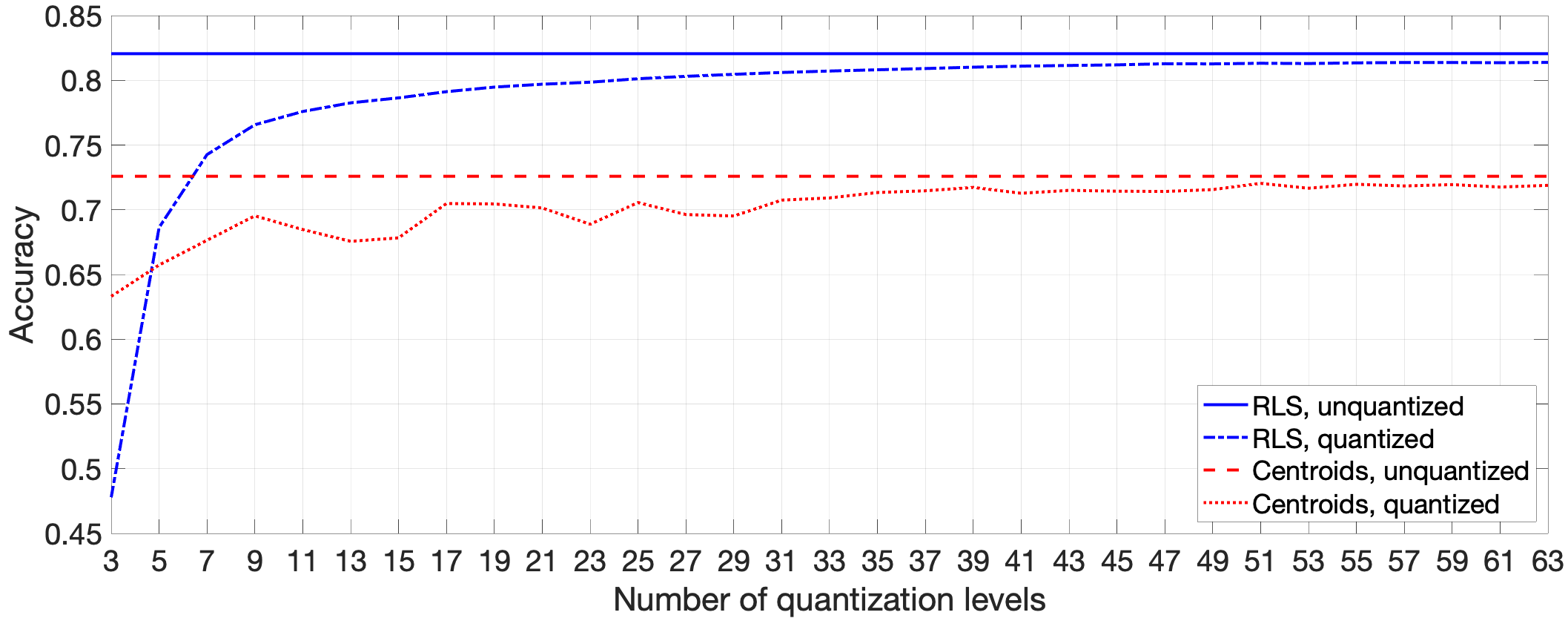}
    \caption{The average accuracy against different number of quantization levels for $\mathbf{W}^{\mathrm{out}}$.
    }
    \label{fig:exp:quant}
\end{figure}

\section{Discussion}


\subsection{Results}

The main point is that, when choosing the compression approach, it is important to consider the burden of the cost  of performing the compression procedure for each local agent, given that the studied solution should be deployed on resource-constrained devices. 
To this end, we note that the proposed compression approach based on HDC/VSA is the most efficient one, being tailored for such devices, while, to a greater extent, the SVD and the lossless ones can be considered as having a lower efficiency but the exact comparison in terms of computational costs is outside the scope of this paper.

\subsection{Relation to other areas}
The proposed approach hinges on principles and premises, which are studied in the federated learning framework~\cite{kairouz2021advances, yang2019federated}. 
Namely, the availability of raw data only at local agents (i.e., ``siloed data'') and the impossibility of sharing such raw data are a standard premise in the federated learning. 
In fact, depending on the definition used, there are certain aspects which might differ; in particular, in our work, there is an absence of a ``master'' agent orchestrating the training, which is often present in the federated learning. This fundamental distinction makes our work challenging in several aspects (computation, combination) that are specific to the fully distributed scenarios. 
Thus, while this work could be considered as a particular case of the federated learning, our work has a significance and purpose by itself and should be studied as a distinct field, given also the different practical application.
Similarly, there are previous studies in  HDC/VSA domain~\cite{KleykoIndustrial2018, ImaniHDColLearn2019}, which investigated the distributed scenario but all of them assumed some elements of centralization.

It also worth noting that while the experiments reported here were done with shallow randomized neural networks, the proposed approach is applicable to deep randomized neural networks~\cite{10.1145/3320060}, which recently gained quite a lot of attraction. 
This is so since the same principle can be used -- only information about the trainable part of the model (i.e., classifier) should be exchanged by the agents, while the randomly chosen part of the network can be the same for every agent. 
In principle, the proposed compression approach could be used even for fully trained neural networks; there is, however, a risk that decompression losses at earlier layers might results in additional errors at the later layers. 
Nevertheless, it is worth clarifying this in the experiments.

The proposed compression approach is conceptually related to the recent idea of using the binding and superposition operations of HDC/VSA to represent parameters of many deep neural networks in a single hypervector~\cite{CheungSuperposition2019, HerscheCompressingBCI2020, ZemanCompressed2021}. 
The difference with the proposed approach is that the classifier is decompressed back to its original shape, which was not the case in~\cite{CheungSuperposition2019}.
The attempts to apply HDC/VSA in the communication domain~\cite{JakimovskiCollective2012, KleykoMACOM2012, KimHDM2018, SimpkinHDWorkflow2019}  are also of relevance but there the goal would usually be to extract the data back from the hypervector without any losses, which is not the case in this work.

\subsection{Quantization of the classifier}
\label{subsec:quant}

In the previous experiments, the weights in $\mathbf{W}^{\mathrm{out}}$ being compressed were real-valued since the proposed compression approach is real-valued as well. 
It is, however, known that neural networks can perform well even when the weights are limited to a few quantization levels~\cite{QuanNN,RVFLFPGA}. 
Therefore, we did an experiment to demonstrate what to potentially expect for the quantization of $\mathbf{W}^{\mathrm{out}}$.
For the sake of simplicity, the experiment was done using the centralized scenario so $N=1$. 
The goal of this experiment was to explore whether it is worth considering quantizing the classifier before the compression. 

Fig.~\ref{fig:exp:quant} presents the average accuracy of the RLS (dash-dotted line) and centroids (dotted line) classifiers (uncompressed)  on all $18$ datasets against the number quantization levels being used. 
The corresponding accuracies (solid and dashed lines) from the unquantized real-valued  $\mathbf{W}^{\mathrm{out}}$ were used as the baselines. 
When $\mathbf{W}^{\mathrm{out}}$ of the RLS classifier was quantized to very few levels ($3$ or $5$) the accuracy was affected significantly. 
It is, however, clear that for the increased number of levels the accuracy of both classifier started to reach their unquantized baselines. 
The results, thus, suggest that allocating one byte per weight should preserve most of the classification performance. 
Note that in the proposed approach, $\mathbf{w}$ with the compressed version of $\mathbf{W}^{\mathrm{out}}$ can also be represented with fewer than $32$ bits per weight. 
It is expected that decreasing the precision of $\mathbf{w}$ will just add some additional noise during the reconstruction. 
Obtaining the quantitative results to characterize the effect of this noise will be a part of the future extension of this work.


\section{Conclusions}

We presented a study exploring the effect of compression on classification performance in a distributed classification scenario.
The main goal was to explore the proposed compression approach, which is based on HDC/VSA principles.
By examining the numerical results obtained on a collection of $18$ datasets for three compression approaches, it is concluded that the proposed approach has a favorable trade-off in terms of accuracy and communication costs.
Also, the finding that small models perform well or on a par with respect to other schemes is worth exploring, and can be considered as enriching the presently relevant discussion regarding the convenience of training large, deep models.
Further, future studies should investigate the sensitivity of the overall performance of the proposed approach with respect to residual coding schemes of matrix coefficients, similarly to what is performed in linear predictive coding and adaptive coding of spectral coefficients, as well as considering some parsimonious model representation strategies as, for instance, Minimum Description Length (MDL) and Bayesian Information Criterion (BIC).

\bibliographystyle{splncs03_unsrt}
\bibliography{references}

\end{document}